\documentclass[runningheads]{llncs}
\usepackage[T1]{fontenc}
\usepackage{graphicx}
\usepackage{bm}
\usepackage{amsfonts}
\usepackage{mathtools}
\usepackage{multirow}
\usepackage{caption}
\usepackage{subcaption}
\usepackage[misc]{ifsym}
\begin{document}
\title{PrUE: Distilling Knowledge from Sparse Teacher Networks}
\toctitle{PrUE: Distilling Knowledge from Sparse Teacher Networks}

\author{Shaopu Wang\inst{1,2}\orcidID{0000-0002-8873-2948} \and
    Xiaojun Chen\textsuperscript{(\Letter)}\inst{2} \and
    Mengzhen Kou\inst{1,2} \and
    Jinqiao Shi\inst{3}
}
\tocauthor{Shaopu~Wang Xiaojun~Chen Mengzhen~Kou Jinqiao~Shi}
\authorrunning{S. Wang et al.}
% First names are abbreviated in the running head.
% If there are more than two authors, 'et al.' is used.
%
\institute{School of Cyber Security, University of Chinese Academy of Sciences, Beijing, China \and
    Institute of Information Engineering, Chinese Academy of Sciences, Beijing, China
    \email{\{wangshaopu,chenxiaojun,koumengzhen\}@iie.ac.cn}\\
    \and
    Beijing University of Posts and Telecommunication, Beijing, China\\
    \email{shijinqiao@bupt.edu.cn}}
\maketitle              % typeset the header of the contribution
\begin{abstract}
    Although deep neural networks have enjoyed remarkable success across a wide variety of tasks, their ever-increasing size also imposes significant overhead on deployment. To compress these models, knowledge distillation was proposed to transfer knowledge from a cumbersome (teacher) network into a lightweight (student) network. However, guidance from a teacher does not always improve the generalization of students, especially when the size gap between student and teacher is large. Previous works argued that it was due to the high certainty of the teacher, resulting in harder labels that were difficult to fit. To soften these labels, we present a pruning method termed Prediction Uncertainty Enlargement (PrUE) to simplify the teacher. Specifically, our method aims to decrease the teacher's certainty about data, thereby generating soft predictions for students. We empirically investigate the effectiveness of the proposed method with experiments on CIFAR-10/100, Tiny-ImageNet, and ImageNet. Results indicate that student networks trained with sparse teachers achieve better performance. Besides, our method allows researchers to distill knowledge from deeper networks to improve students further. Our code is made public at: \url{https://github.com/wangshaopu/prue}.

\end{abstract}
\section{Introduction}\label{intro}
Neural networks have gained remarkable practical success in many fields~\cite{gpt32020}. In practice, researchers usually introduce more layers and parameters to make the network deeper~\cite{zagoruyko2016wide} and wider~\cite{ioffe2015batch} for achieving better performance. However, these over-parameterized models also incur huge computational and storage overhead~\cite{journeybegin2018}, which makes deploying them on edge devices impractical. Therefore, several methods have been proposed to shrink neural networks, e.g., network pruning~\cite{brain1989,pruning2015}, quantization~\cite{quantization2014}, and knowledge distillation~\cite{knowledgedistillatio2015}. Among these approaches, knowledge distillation has been widely utilized in many fields~\cite{largescale2018,fedgen2021}. Generally speaking, it utilizes a pre-trained teacher to produce supervision for students. In this way, a lightweight student network can achieve similar generalization as the teacher.

\begin{table}[tbp]
    \caption{The test accuracy in percentage of various teachers and ResNet-8 as the student.}\label{tab1_ls}
    \centering
    \begin{tabular}{|c|c|c|c|c|}
        \hline
        \multicolumn{1}{|l|}{} & ResNet-8         & ResNet-20 w/o LS & ResNet-20 w/ LS  & ResNet-32 w/o LS \\ \hline
        Teacher Acc.           & 87.56($\pm0.20$) & 91.72($\pm0.21$) & 92.06($\pm0.26$) & 92.99($\pm0.12$) \\
        Student Acc.           & -                & 88.05($\pm0.18$) & 86.13($\pm0.22$) & 87.60($\pm0.08$) \\ \hline
    \end{tabular}
\end{table}

\begin{figure}[tbp]
    \centering
    \includegraphics[width=\textwidth]{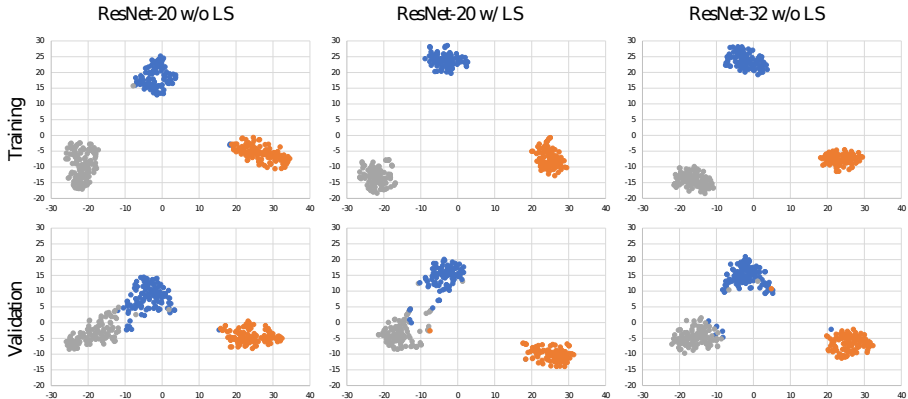}
    \caption{Visualization of network predictions. We randomly select some training samples from the three classes of CIFAR-10 ``airplane" (gray), ``automobile" (blue), and ``bird" (yellow), and then perform t-SNE dimensionality reduction~\cite{van2008visualizing} on network predictions. Note that the x-y axis has no real meaning here.} \label{fig1_ls}
\end{figure}

Although this paradigm of encouraging students to mimic teachers' behaviors has proven to be a promising way, some recent works~\cite{whenhelp2019,doeswork2021} argued that knowledge distillation is not always effective. Specifically, it is found that well-behaved teachers failed to improve student generalization under certain circumstances. For instance, Müller et al. \cite{whenhelp2019} discovered that teachers pre-trained with label smoothing (LS)~\cite{szegedy2016rethinking}, a commonly used technique to regularize models, will distill inferior students, even though the teacher's generalization has been improved. They attribute this phenomenon to the fact that LS tends to erase the relative information within a class. As a result, teachers generate harder labels that are difficult for students to fit. Meanwhile, Mirzadeh et al.~\cite{teacherassistant2020} investigated another more common scenario. When there exists a large capacity gap between students and teachers, the former will perform worse. Coincidentally, their experiments lead to a similar conclusion that well-performed teachers fail to generate soft targets.

To investigate the relationship between network capacity and label smoothing, we train ResNet-20 and ResNet-32 on CIFAR-10 and report the results of visualizing their predictions for the classes ``airplane", ``automobile" and ``bird" in Fig.~\ref{fig1_ls}. The first row represents examples from the training set, while the second row is from the validation set. As revealed in the first column, a ResNet-20 trained without label smoothing (w/o LS) produces predictions scattered in some broad clusters. We also notice that blue dots (automobile) and gray dots (airplanes) in the validation set tend to be mixed at the boundary. A possible explanation is that these vehicles are more similar in some features than the yellow dots (birds), and it causes some misclassification. While in the second column, a ResNet-20 is trained with a label smoothing factor of 0.2. We observe that LS encourages samples in the training set to be equidistant from other classes' centers. What is striking in this figure is the third column. We train a ResNet-32 and notice that it acts in a similar pattern to LS. They both compact each class cluster. Next, we use ResNet-8 as a student to validate the effectiveness of knowledge distillation. The accuracy results, as shown in Table~\ref{tab1_ls}, confirm that while label smoothing and network deepening can improve the teacher network, they will degrade the generalization of students as expected.

A possible speculation is that although the generalization of the networks can be improved by the above two measures, their uncertainty about the data is also reduced. As a result, teachers tend to produce similar overconfident predictions for all intra-class samples and distill inferior students. In this work, we propose to improve knowledge distillation by increase teachers' uncertainty. Fortunately, a statistical metric, which we term prediction uncertainty, has been proposed by~\cite{truly2021} to quantify this phenomenon. Following this work, we propose a criterion to identify the effect of weights on uncertainty in the teacher network. Then we prune those less-contributing weights before distillation. Differing from traditional pruning algorithms that focus on generalization, our method aims to reduce the generalization error of student networks by softening teacher predictions. We name our method Prediction Uncertainty Enlargement (PrUE).

We evaluate our pruning method on CIFAR-10/100, Tiny-ImageNet, and ImageNet classification datasets with some modern neural networks. Specifically, we first verify that label smoothing and network deepening reduce generalization error with a sacrifice of prediction uncertainty. The following distillation experiments show a positive correlation between the student's accuracy and the teacher's prediction uncertainty. However, the teacher's accuracy does not play a crucial role in knowledge distillation. Generally, large networks struggle to distill stronger students despite their high accuracy. To bridge this gap, we apply PrUE to the aforementioned teacher networks and distill their knowledge to students. Results show that our method can increase the teacher's prediction uncertainty, resulting in better performance improvement for students than existing distillation methods. We also compare PrUE with several other pruning schemes and observe that sparse teacher networks distill good students, but PrUE usually presents better performance.

\textbf{Contributions:} our contributions in this paper are as follows.
\begin{itemize}
    \item We empirically investigate the impact of label smoothing and network capacity on knowledge distillation. Interestingly, They both prevent teachers from generating soft labels and impair knowledge distillation, despite the improved accuracy of teachers themselves.
    \item We apply a statistical metric to quantify the softness of labels. Based on this, PrUE is proposed to increase the teacher's prediction uncertainty.
    \item We perform experiments on CIFAR-10/100, Tiny-ImageNet, and ImageNet with widely varying CNN networks. Results suggest that sparse teacher networks usually distill better students than dense ones. Besides, PrUE outperforms existing distillation and pruning schemes.
\end{itemize}

\section{Related Work}
\subsubsection{Network Pruning.} The motivation behind network pruning is that there is a mass of redundant parameters in the neural network~\cite{denil2013predicting}. Previous works have demonstrated that these parameters can be removed safely. Therefore, Lecun et al.~\cite{brain1989} proposed removing parameters in an unstructured way by calculating the Hessian of the loss with respect to the weights. Furthermore, Han et al.~\cite{pruning2015} proposed a magnitude-based pruning method to remove all weights below s predefined threshold. Recently, Frankle et al.~\cite{lth2019} proposed the ``Lottery Ticker Hypothesis" that there exist sparse subnetworks that, when trained in isolation, can reach test accuracy comparable to the original network. Furthermore, Miao et al.~\cite{miao2021learning} proposed a framework that can prune neural networks to any sparsity ratio with only one training.

\subsubsection{Soft Labels.} Theoretically, the widely used one-hot labels could lead to overfitting. Therefore, label smoothing was proposed to generate soft labels, thereby delivering regularization effects. On the other hand, there were usually some noisy labels in the dataset that mislead deep learning models, and a recent work~\cite{lukasik2020does} noted that label smoothing could help mitigate label noise. However, label smoothing could only add random noise and cannot reflect the relationship between labels. Another well-known paradigm for generating soft labels was knowledge distillation~\cite{knowledgedistillatio2015}. Differing label smoothing, knowledge distillation required a pretrained teacher to produce soft labels for each training example. Therefore, Yuan et al.~\cite{yuan2019revisit} regarded it as a dynamic form of label smoothing. Although the original distillation scheme focused on transferring dark knowledge from large to small models, Zhang et al.~\cite{dml2018} had found that these generated soft labels can be used for distributed machine learning. Therefore, some recent works~\cite{largescale2018,fedgen2021} proposed distillation-based communication schemes to save bandwidth.

\subsubsection{Pruning In Distillation.} Both network pruning and knowledge distillation are widely used model compression methods. Therefore, some recent works proposed combining them together to achieve higher compression ratios. For instance, Xie et al.~\cite{xie2021model} used this paradigm to customize a compression scheme for the identification of Person re-identification (ReID). Chen et al.~\cite{chen2020learning} proposed to use pruning and knowledge distillation to train a lightweight detection model, to achieve synthetic aperture radar ship real-time detection at a lower cost. Meanwhile, Aghli et al.~\cite{aghli2021combining} introduced a compression scheme of convolutional neural networks, mainly exploring how to combine pruning and knowledge distillation methods to reduce the scale of ResNet with the guarantee of accuracy. Neill et al.~\cite{neill2021deep} proposed a pruning-based self-distillation scheme using distillation as the pruning criterion to maximize the similarity of network representations before and after pruning. Cui et al.~\cite{cui2021joint} proposed a joint model compression method that combines structured pruning and dense knowledge distillation. However, these researches focused on simplifying student networks. In fact, they amplify the capacity gap between students and teachers.

\section{Background}
Producing soft labels has been shown to be an effective regularizer. In practice, encouraging networks to fit soft labels prevents overfitting. In this section, we introduce a statistical metric quantifying label softness.
\subsection{Preliminaries}
\subsubsection{Notations.} Given a $K$-class classification task, We denote by $\mathcal{D}$ the training dataset, consisting of $m$ i.i.d tuples $\{(\bm{x}_1, \bm{y}_1), \ldots, (\bm{x}_m, \bm{y}_m)\}$ where $\bm{x}_i \in \mathbb{R}^{d\times 1}$ is the input data and $\bm{y}_i\in\{0, 1\}^K$ is the corresponding one-hot class label. Let $\bm{y}[i]$ be the $i$-th element in $\bm{y}$, and $\bm{y}[c]$ is 1 for the ground-truth class and 0 for others.

\subsubsection{Knowledge Distillation.} For a teacher network $f(\bm{w}_\mathcal{T})$ parameterized by $\bm{w}_\mathcal{T}$, let $a(\bm{w}_\mathcal{T}, \bm{x}_i)$ and $f(\bm{w}_\mathcal{T}, \bm{x}_i)$ correspond to its logits and prediction for $\bm{x}_i$, respectively. In vanilla supervised learning, $f(\bm{w}_\mathcal{T})$ is usually trained on $\mathcal{D}$ with cross-entropy loss
\begin{equation}
    \mathcal{L}_{CE}=-\sum_{i=1}^{m}\bm{y}_i\log f(\bm{w}_\mathcal{T}, \bm{x}_i)
\end{equation}
where $f(\bm{w}_\mathcal{T}, \bm{x}_i)=softmax(a(\bm{w}_\mathcal{T}, \bm{x}_i))$.

As for a student network $f(\bm{w}_\mathcal{S})$, its logits and prediction for $\bm{x}_i$ are denoted as $a(\bm{w}_\mathcal{S}, \bm{x}_i)$ and $f(\bm{w}_\mathcal{S}, \bm{x}_i)$. In knowledge distillation, $f(\bm{w}_\mathcal{S})$ is usually trained with a given temperature $\tau$ and KL-divergence loss
\begin{equation}
    \mathcal{L}_{KD}=-\sum_{i=1}^{m}\tau^2KL(a(\bm{w}_\mathcal{T}, \bm{x}_i), a(\bm{w}_\mathcal{S}, \bm{x}_i))
\end{equation}

When the hyperparameter $\tau$ is set to 1, we can regard the distillation process as training $f(\bm{w}_\mathcal{S})$ on a new dataset $\{(\bm{x}_1, f(\bm{w}_\mathcal{T}, \bm{x}_1), \ldots, (\bm{x}_m, f(\bm{w}_\mathcal{T}, \bm{x}_m)\}$ with soft labels provided by a teacher. The key idea behind knowledge distillation is to encourage the student $f(w_\mathcal{S})$ to mimic the behavior of the teacher $f(w_\mathcal{T})$. In practice, researchers usually use correct labels to improve soft labels, especially when the generalization of teachers is poor. Therefore, the practical loss function for the student is modified as follows:
\begin{equation}
    \mathcal{L}_{student}=\sum_{i=1}^{m}(1-\lambda)\mathcal{L}_{CE}+\lambda\mathcal{L}_{KD}
\end{equation}
where $\lambda$ is another hyperparameter that controls the trade-off between the two losses. We refer to this approach as \textit{Logits($\tau$)} through the paper.

\subsubsection{Label Smoothing.} Similar to knowledge distillation, label smoothing aims to replace hard labels to penalize overfitting. Instead, it does not involve a teacher network. Specifically, label smoothing modifies one-hot hard label vector $\bm{y}$ with a mixture of weighted origin $\bm{y}$ and a uniform distribution:
\begin{equation}
    \bm{y}_c=\begin{cases}1-\alpha       & \textnormal{if}\; c=label, \\
             \alpha / (K-1) & \textnormal{otherwise}.
    \end{cases}
\end{equation}
where $\alpha\in[0, 1]$ is the hyperparameter flattening the one-hot labels.

Label smoothing has been a widely used \textit{trick} to improve network generalization. A prior work~\cite{truly2021} observes that although the network trained with label smoothing suffers a higher cross-entropy loss on the validation set, its accuracy is better than that without label smoothing.

\subsection{Prediction Uncertainty}
To observe the effect of label smoothing on the penultimate layer representations, Müller et al.~\cite{whenhelp2019} proposed a visualization scheme based on squared Euclidean distance. Similarly, we use t-SNE in Section~\ref{intro} to visualize the predictions. However, we cannot conduct numerical analysis on these intuitive presentations. To further measure the label softness quantitatively and address the \textit{erasing} phenomenon caused by label smoothing, Shen et al.~\cite{truly2021} propose a simple yet effective metric. The definition is as follows\footnote{It was called \textit{stability} in the origin paper. We modify it for the purpose of our work.} :
\begin{equation}
    \delta(\bm{w}) = \frac{1}{K}\sum_{c=1}^{K}(\frac{1}{\bm{n}_c}\sum_{i=1}^{\bm{n}_c}\Vert f(\bm{w}, \bm{x}_i)[c] - \tilde{f}(\bm{w}, \bm{x}_i)[c]\Vert^2)
\end{equation}
where class $c$ contains $\bm{n}_c$ samples. $\tilde{f}(\bm{x}_i)[c]$ is the mean of in $f(\bm{x}_i)$ class c. The key idea behind this metric is to use the variance of intra-class probabilities to measure the uncertainty of network predictions.

Now we discuss how prediction uncertainty influences knowledge distillation. Assume an ideal network classifies each input precisely, and it is absolutely certain of each prediction. Correspondingly, this network is commonly regarded as a perfect model that achieves excellent generalization and low loss on the validation set. However, it tends to produce one-hot labels that fail to inform student networks about the similarity between classes, i.e., dark knowledge. At this point, the certainty of the teacher network downgrades the knowledge distillation to vanilla training. Applying label smoothing to the distillation process could help to moderate the teacher's overconfidence. Unfortunately, this trick merely tells students that airplanes and birds have the same probability as automobiles. Therefore, we aim to make teachers feel uncertain between the automobile and the airplane, thus improving the generalization behavior of the student network.

We next work on simplifying the teacher network to enlarge its prediction uncertainty. Specifically,  we utilize network pruning to close the capacity gap between teachers and students.

\section{Prediction Uncertainty Enlargement}
In deep model compression, network pruning aims to deliver the regularization effect to neural networks by simply removing parameters. Following the discussion above, we introduce auxiliary indicator variables $\bm{m}\in\{0,1\}^l$ representing the pruning mask. Then the enlargement of prediction uncertainty is formulated as an optimization problem as:
\begin{equation}
    \begin{aligned}
        \label{binary_op}
        \mathop{\max}\limits_{\bm{m}}\delta(\bm{m}\odot \bm{w})= & \mathop{\max}\limits_{\bm{m}}\frac{1}{K}\sum_{c=1}^{K}(\frac{1}{\bm{n}_c}\sum_{i=1}^{\bm{n}_c}\Vert f(\bm{m}\odot\bm{w}, \bm{x}_i)[c] - \tilde{f}(\bm{m}\odot\bm{w}, \bm{x}_i)[c]\Vert^2), \\
        s.t.                                                     & \quad \bm{m}\in\{0, 1\}^l, \quad \Vert \bm{m} \Vert_0\leq s,
    \end{aligned}
\end{equation}
where $\odot$ denotes the Hadamard product.

Solving such a combinatorial optimization problem requires computing its $\delta(\bm{m}\odot \bm{w})$ for each candidate in the solution space, that is, it requires up to $l\times l$ forward passes over the training dataset. However, the number of network parameters has increased substantially recently. Since an arms race of training large models has begun, millions of calculations $\delta(\bm{m}\odot \bm{w})$ are unacceptable.

Following~\cite{koh2017understanding,snip2019}, we next measure the impact of each weight on the network uncertainty and then prune less-contributing weights greedily. Since it is impractical to directly solve this optimization problem with respect to binary variables $\bm{m}$, we first relax $\bm{m}$ into real variables $\bm{m}\in[0,1]^l$. This change can be seen as a form of soft pruning, where the corresponding mask $\bm{m}[j]$ is gradually reduced from 1 to 0. In this way, the optimization problem is differentiable with respect to $\bm{m}$. We rewrite Optimization~(\ref{binary_op}) as follows:

\begin{equation}
    \begin{aligned}
        \label{real_op}
        \mathop{\max}\limits_{\bm{m}}\delta(\bm{m}\odot \bm{w})= & \mathop{\max}\limits_{\bm{m}}\frac{1}{K}\sum_{c=1}^{K}(\frac{1}{\bm{n}_c}\sum_{i=1}^{\bm{n}_c}\Vert f(\bm{m}\odot\bm{w}, \bm{x}_i)[c] - \tilde{f}(\bm{m}\odot\bm{w}, \bm{x}_i)[c]\Vert^2), \\
        s.t.                                                     & \quad \bm{m}\in[0, 1]^l, \quad \Vert \bm{m} \Vert_0\leq s,
    \end{aligned}
\end{equation}

This modification allows us to perturb the mask instead of setting it to zero. For the weight $\bm{w}[j]$, we add an infinitesimal perturbation $\epsilon$ to the mask $\bm{m}[j]$ to obtain its influence on $\delta(\bm{m}\odot \bm{w})$. Its magnitude of differential $\triangle\delta_j(\bm{m}\odot \bm{w})$ indicates the dependence of $\delta(\bm{m}\odot \bm{w})$ on $\bm{w}[j]$. Next, we find the derivative of $\delta(\bm{m}\odot \bm{w})$ with respect to $\bm{m}[j]$ as follows:
\begin{equation}
    \begin{aligned}
        \label{real_effect}
        \lim_{\epsilon\to 0} \frac{\delta(\bm{m}\odot\bm{w})- \delta((\bm{1}-\epsilon\bm{e}_j)\bm{m}\odot\bm{w})}{\epsilon} = \lim_{\epsilon\to 0} \dfrac{\partial \delta(\bm{m}\odot \bm{w})}{\partial \bm{m}[j]}=g_j(\bm{w}).
    \end{aligned}
\end{equation}
where $\bm{e}_j$ is a one-hot vector $[0, ..., 0, 1, 0,..., 0]$ with a 1 at position $j$.

Thus, we measure the importance of the weight $\bm{w}[j]$ to the prediction uncertainty. To this end, we regard $\vert g_j(\bm{w}) \vert$ as the proposed criterion. Given a desired sparsity $s$, we can achieve prediction uncertainty enlargement by pruning $s\times l$ weights that contribute less to the variance.

The key to our approach is to find the derivative of the uncertainty with respect to the pruning mask of each weight. However, restricted by the modern computing device, PrUE still faces some practical problems. Note that Optimization~(\ref{real_op}) calls $f(\bm{w})$ twice, which requires the automatic differentiation algorithm to perform two forward-backward pass through the computational graph. Modern deep learning frameworks like PyTorch usually free gradient tensors after the first backward pass to save memory. That is, our method consumes more resources due to retaining the computational graph.

On the other hand, our method requires computing the averaged intra-class probabilities for each class. In practice, researchers typically perform stochastic gradient descent by randomly selecting a mini-batch of training data, where the batchsize ranges from 128 to 1024. For a 10-class classification task like CIFAR-10, this batchsize is sufficient to estimate $\tilde{f}(\bm{x})[c]$, while not for ImageNet-1k containing 1000 classes. In fact, most classes in ImageNet-1k only appear once or twice in a batch, making accurate estimation of $\tilde{f}(\bm{x})[c]$ impractical.

One could take straightforward measures such as saving intermediate values of the graph or leveraging more devices, but this would result in additional overhead. Instead, we employ a simple yet effective trick to decompose the optimization into two steps. Specifically, we first compute $\tilde{f}(\bm{x})[c]$ for each class with the computational graph detached, then sort the dataset by labels, thus guaranteeing that only class $c$ appears in each batch. Finally, ${f}(\bm{x})[c]$ can be estimated in the current batch. We empirically observed that this trick only slightly affects the results, but saves appreciable memory.

\section{Experiments}
In this section, we empirically investigate the effect of our proposed method on knowledge distillation. In addition, we compare PrUE with other distillation and pruning methods. The results show that our paradigm of distilling knowledge from sparse teacher networks tends to yield better students. Moreover, PrUE can exhibit better performance.

\begin{table}[t]
    \caption{Number of weights and training hyperparameters in our experiments.} \label{hyper}\centering
    \begin{tabular}{|c|c|c|c|c|c|}
        \hline
        Dataset                        & Network                                       & \# Weights & Epochs               & Batchsize            & Schedule                                                                           \\ \hline
        \multirow{5}{*}{CIFAR-10}      & ResNet-8~\cite{resnet2016}                    & 78K        & \multirow{5}{*}{160} & \multirow{5}{*}{512} & \multirow{5}{*}{\begin{tabular}[c]{@{}c@{}}10x drop at \\ 81, 122\end{tabular}}    \\
                                       & ResNet-20                                     & 272K       &                      &                      &                                                                                    \\
                                       & ResNet-32                                     & 466K       &                      &                      &                                                                                    \\
                                       & ResNet-56                                     & 855K       &                      &                      &                                                                                    \\
                                       & ResNet-110                                    & 1.7M       &                      &                      &                                                                                    \\ \hline
        \multirow{5}{*}{CIFAR-100}     & ResNet-8                                      & 83.9K      & \multirow{5}{*}{160} & \multirow{5}{*}{512} & \multirow{5}{*}{\begin{tabular}[c]{@{}c@{}}10x drop at \\ 81, 122\end{tabular}}    \\
                                       & ResNet-20                                     & 278K       &                      &                      &                                                                                    \\
                                       & ResNet-32                                     & 472K       &                      &                      &                                                                                    \\
                                       & ResNet-56                                     & 861K       &                      &                      &                                                                                    \\
                                       & ResNet-110                                    & 1.7M       &                      &                      &                                                                                    \\ \hline
        \multirow{2}{*}{Tiny-ImageNet} & ResNext-50~\cite{xie2017aggregated}           & 15.0M      & \multirow{2}{*}{200} & \multirow{2}{*}{256} & \multirow{2}{*}{\begin{tabular}[c]{@{}c@{}}5x drop at\\ 60, 120, 160\end{tabular}} \\
                                       & ShuffleNet V2~\cite{shuffle2018}              & 1.5M       &                      &                      &                                                                                    \\ \hline
        \multirow{2}{*}{ImageNet}      & EffcientNet-B2~\cite{efficient2019}           & 9.1M       & \multirow{2}{*}{90}  & \multirow{2}{*}{256} & \multirow{2}{*}{\begin{tabular}[c]{@{}c@{}}10x drop at\\ 30, 60\end{tabular}}      \\
                                       & MobileNet V3 Small~\cite{howard2019searching} & 2.5M       &                      &                      &                                                                                    \\ \hline
    \end{tabular}
\end{table}

\subsubsection{Implementation Details.} We conduct all experiments on 8 * NVIDIA Tesla A100 GPU. The sparsity level is defined to be $s=k/ l\times 100(\%)$, where $k$ is the number of zero weights, and $l$ is the total number of network weights. All networks are trained with SGD with Nesterov momentum. We set the initial learning rate to 0.1, momentum to 0.9. Table~\ref{hyper} describes the number of parameters of all the networks and corresponding training hyperparameters. During distillation, we set $\lambda$ to 1 for CIFAR-10 and 0.1 for the rest tasks.

\subsection{The Effect of LS on Knowledge Distillation}

We first investigate the compatibility of label smoothing and knowledge distillation on CIFAR-10 and CIFAR-100. Specifically, we train ResNet-20/32/56/100 with label smoothing turned on or turned off, then distill their knowledge into ResNet-8. Table~\ref{ls_distill} presents the accuracy of student networks supervised by various teachers. We also report the vanilla supervised training results of ResNet-8 for baseline comparison.

\begin{table}[htbp]
    \centering
    \caption{The test accuracy of a fixed student with various teachers trained without (w/o) or with (w/) label smoothing. The vanilla supervised results of ResNet-8 is also reported.}
    \label{ls_distill}
    \begin{tabular}{|c|c|c|cccc|}
        \hline                     & Vanilla                &       & \multicolumn{1}{c|}{ResNet-20} & \multicolumn{1}{c|}{ResNet-32} & \multicolumn{1}{c|}{ResNet-56} & ResNet-110       \\ \hline
        \multirow{2}{*}{CIFAR-10}  & \multirow{2}{*}{87.56} & w/ LS & 86.62($\pm0.21$)               & 85.56($\pm0.25$)               & 85.61($\pm0.03$)               & 85.88($\pm0.19$) \\ \cline{3-7}   &     & w/o LS & 88.36($\pm0.12$)    & 87.48($\pm0.22$)    & 87.50($\pm0.10$)  & 87.47($\pm0.17$)  \\ \hline
        \multirow{2}{*}{CIFAR-100} & \multirow{2}{*}{59.36} & w/ LS & 58.75($\pm0.23$)               & 58.73($\pm0.16$)               & 59.14($\pm0.14$)               & 58.52($\pm0.25$) \\ \cline{3-7}    &      & w/o LS & 59.81($\pm0.19$)    & 59.50($\pm0.24$)    & 59.47($\pm0.17$)   & 59.76($\pm0.09$)   \\ \hline
    \end{tabular}
\end{table}

\begin{figure}[htbp]
    \centering
    \includegraphics[width=0.8\textwidth]{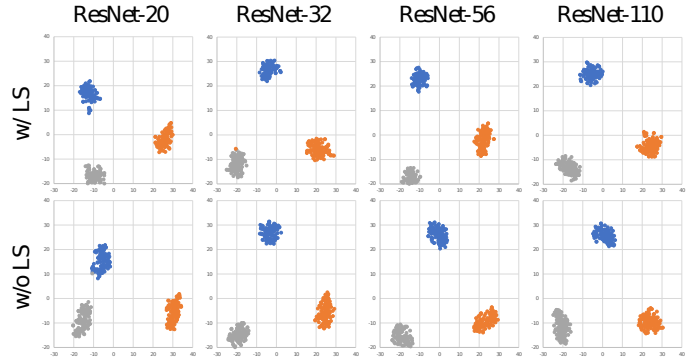}
    \caption{Visualization of predictions of more network structures.} \label{fig2_lsscatter}
\end{figure}

Although deep neural networks are well known for their generalization ability, they fail to bring proportional improvement for students. In particular, ResNet-20 tends to distill better students than other well-generalized teachers. Similarly, teachers trained with hard labels achieve better distillation results compared to those trained with label smoothing. To demonstrate this phenomenon, we provide visualizations of these teachers' predictions in Fig.~\ref{fig2_lsscatter}. As we can see, network deepening and label smoothing compacts each cluster and thus impairs knowledge distillation in Table~\ref{ls_distill}.

\subsection{Comparison with Other Distillation Methods}

Intuitively, improved teachers are overconfident in each sample, thus producing harder predictions containing low information. To enlarge teacher uncertainty without sacrificing generalization, we apply PrUE to prune them, and then fine-tune them to restore accuracy.

\begin{figure}[htbp]
    \centering
    \includegraphics[width=0.8\textwidth]{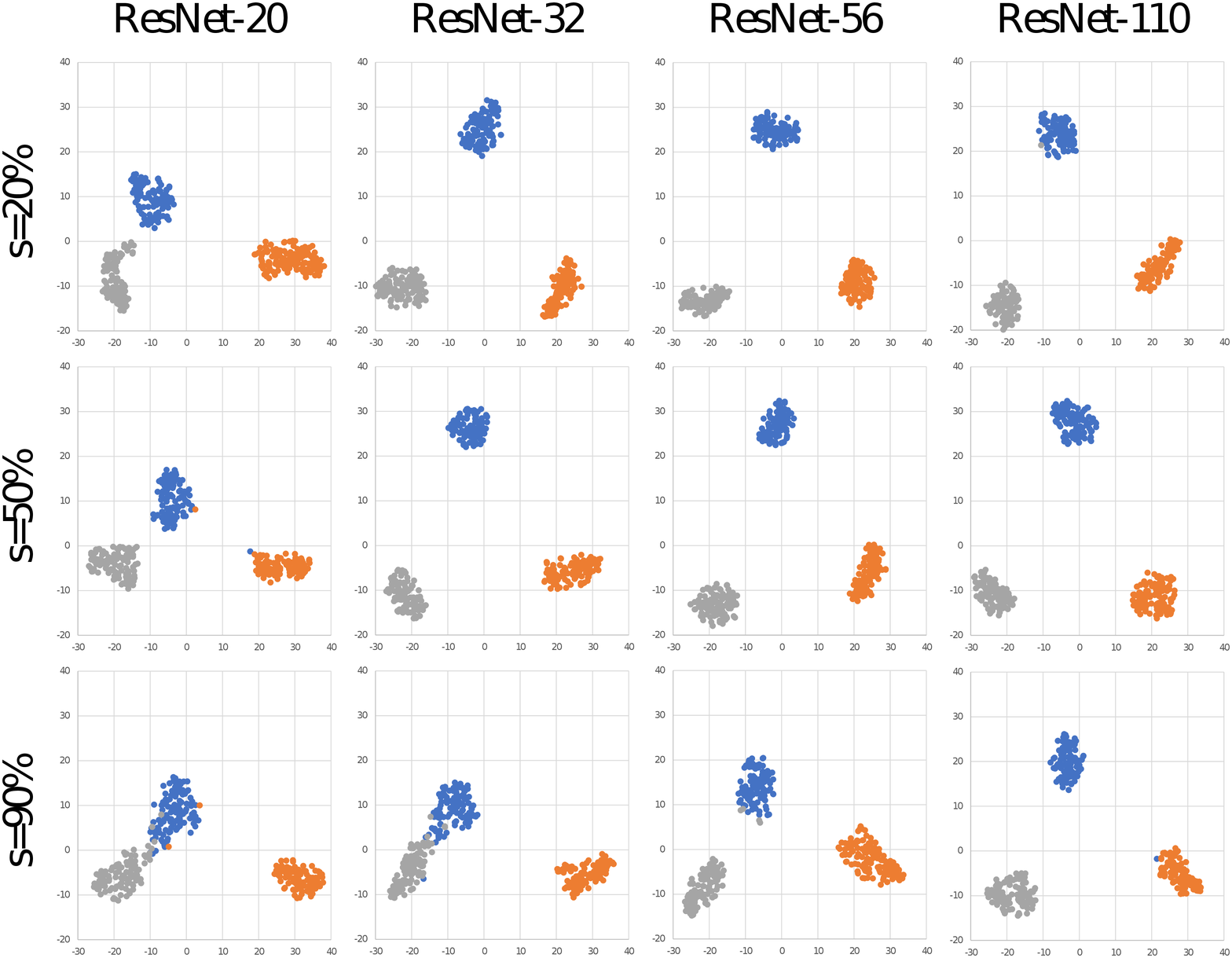}
    \caption{Predictive visualization of networks with varying sparsity $s$. As the network deepens, the predictions get tighter. While the increasing sparsity spreads the predictions into broad clusters.} \label{fig3_prunescatter}
\end{figure}

\begin{table}[htbp]
    \centering
    \caption{The test accuracy (\%) and uncertainty (1e-2) of teacher networks with varying sparsity.}
    \label{teacher_property}
    \begin{tabular}{|c|c|cc|cc|cc|cc|}
        \hline
        \multirow{2}{*}{Dataset}   & \multirow{2}{*}{Sparsity} & \multicolumn{2}{c|}{ResNet-20} & \multicolumn{2}{c|}{ResNet-32} & \multicolumn{2}{c|}{ResNet-56} & \multicolumn{2}{c|}{ResNet-110}                                                                           \\ \cline{3-10}
                                   &                           & \multicolumn{1}{c|}{Acc.}      & Uncer.                         & \multicolumn{1}{c|}{Acc.}      & Uncer.                          & \multicolumn{1}{c|}{Acc.} & Uncer. & \multicolumn{1}{c|}{Acc.} & Uncer. \\ \hline
        \multirow{4}{*}{CIFAR-10}  & $s=0$                     & 91.72                          & 6.40                           & 93.17                          & 3.34                            & 93.40                     & 2.04   & 93.38                     & 1.55   \\
                                   & $s=20\%$                  & 92.82                          & 6.12                           & 93.54                          & 3.19                            & 93.75                     & 2.12   & 94.14                     & 1.31   \\
                                   & $s=50\%$                  & 91.97                          & 8.19                           & 93.08                          & 4.23                            & 93.77                     & 2.39   & 93.75                     & 1.74   \\
                                   & $s=90\%$                  & 87.98                          & 22.87                          & 90.63                          & 17.49                           & 91.64                     & 13.00  & 92.13                     & 7.28   \\ \hline
        \multirow{4}{*}{CIFAR-100} & $s=0$                     & 68.61                          & 26.01                          & 69.65                          & 19.16                           & 71.29                     & 10.72  & 71.84                     & 5.72   \\
                                   & $s=20\%$                  & 69.04                          & 25.83                          & 70.44                          & 18.77                           & 72.01                     & 10.47  & 73.36                     & 5.55   \\
                                   & $s=50\%$                  & 68.26                          & 27.97                          & 69.18                          & 22.55                           & 71.33                     & 14.56  & 72.83                     & 7.18   \\
                                   & $s=90\%$                  & 54.30                          & 28.33                          & 60.49                          & 30.53                           & 62.92                     & 30.78  & 62.16                     & 30.52  \\ \hline
    \end{tabular}
\end{table}

Fig.~\ref{fig3_prunescatter} visualizes these sparse teacher networks. As the sparsity $s$ increases, the teacher's predictions are scattered into wider clusters. We also observe that a higher sparsity is appropriate for deep networks such as ResNet-110. On the other hand, Table~\ref{teacher_property} provides quantitative results. It suggests that PrUE can effectively improve teachers' uncertainty with slight loss in performance.

\begin{table}[htbp]
    \centering
    \caption{The test accuracy of ResNet-8 on CIFAR-10 using different distillation methods. TA(20), TA(32) refers to using ResNet-20 and ResNet-32 as a teacher assistant, respectively. }\label{cifar10_results}
    \begin{tabular}{|c|cccc|}
        \hline
        \multirow{2}{*}{}                               & \multicolumn{4}{c|}{CIFAR-10}                                                                                                \\ \cline{2-5}
                                                        & \multicolumn{1}{c|}{ResNet-20} & \multicolumn{1}{c|}{ResNet-32} & \multicolumn{1}{c|}{ResNet-56} & ResNet-110                \\ \hline
        Logit($\tau=1$)~\cite{knowledgedistillatio2015} & 88.36($\pm0.16$)               & 87.48($\pm0.22$)               & 87.50($\pm0.29$)               & 87.47($\pm0.28$)          \\ \cline{1-1}
        Logit($\tau=4$)                                 & 88.72($\pm0.26$)               & 88.39($\pm0.17$)               & 88.66($\pm0.21$)               & 88.34($\pm0.29$)          \\ \cline{1-1}
        FitNet~\cite{fitnet2014}                        & 87.00($\pm0.24$)               & 86.83($\pm0.27$)               & 86.68($\pm0.08$)               & 86.62($\pm0.10$)          \\ \cline{1-1}
        AT~\cite{zagoruyko2016paying}                   & 86.64($\pm0.14$)               & 86.37($\pm0.16$)               & 86.71($\pm0.09$)               & 86.76($\pm0.17$)          \\ \cline{1-1}
        PKT~\cite{passalis2018learning}                 & 87.41($\pm0.04$)               & 87.30($\pm0.24$)               & 87.26($\pm0.13$)               & 87.08($\pm0.19$)          \\ \cline{1-1}
        TA(20)~\cite{teacherassistant2020}              & -                              & 87.55($\pm0.26$)               & 87.87($\pm0.19$)               & 87.66($\pm0.16$)          \\ \cline{1-1}
        TA(32)                                          & -                              & -                              & 87.83($\pm0.10$)               & 87.37($\pm0.28$)          \\ \hline
        PrUE($s=20\%$)                                  & 88.89($\pm0.11$)               & 88.30($\pm0.06$)               & 88.49($\pm0.19$)               & 88.47($\pm0.26$)          \\ \cline{1-1}
        PrUE($s=50\%$)                                  & \textbf{89.17}($\pm0.19$)      & \textbf{88.39}($\pm0.07$)      & 88.68($\pm0.23$)               & 89.22($\pm0.15$)          \\ \cline{1-1}
        PrUE($s=90\%$)                                  & 87.01($\pm0.20$)               & 87.95($\pm0.24$)               & \textbf{89.08}($\pm0.24$)      & \textbf{89.27}($\pm0.18$) \\ \hline
    \end{tabular}
\end{table}

Next, we distill knowledge from these sparse teachers to a ResNet-8. Meanwhile, we compare our method with other distillation methods. Table~\ref{cifar10_results} and Table~\ref{cifar100_results} depicts the results of students performance on CIFAR-10 and CIFAR-100, respectively. It is worth noting that $\lambda$ is set to 0 on CIFAR-10, which means that our method can only obtain the teacher's prediction, while the others can receive the ground truth. Although this is an unfair comparison, PrUE still outperforms existing distillation methods notably. Another interesting observation is that teachers with high uncertainty distill better students, even when their accuracy is hurt by pruning. Therefore, we conclude that teacher uncertainty plays an important role in knowledge distillation, rather than accuracy.

\subsection{Comparison with Other Pruning Methods}
With promising results on distillation, we further compare PrUE with other pruning methods. In particular, we first train the teacher from scratch and apply several one-shot pruning algorithms (Magnitude~\cite{pruning2015,liu2021lottery}, SNIP~\cite{snip2019}, Random~\cite{whymiss2021}, PrUE) to remove a portion of weights of the trained network, then fine-tune these pruned networks until convergence. We use ResNet-8 as a student to evaluate the distillation performance of these sparse teachers.

\begin{table}[htbp]
    \centering
    \caption{The test accuracy of ResNet-8 on CIFAR-100 using different distillation methods.}\label{cifar100_results}
    \begin{tabular}{|c|cccc|}
        \hline
        \multirow{2}{*}{} & \multicolumn{4}{c|}{CIFAR-100}                                                                                               \\ \cline{2-5}
                          & \multicolumn{1}{c|}{ResNet-20} & \multicolumn{1}{c|}{ResNet-32} & \multicolumn{1}{c|}{ResNet-56} & ResNet-110                \\ \hline
        Logit($\tau=1$)   & 59.51($\pm0.10$)               & 59.25($\pm0.23$)               & 59.09($\pm0.08$)               & 59.56($\pm0.26$)          \\ \cline{1-1}
        Logit($\tau=4$)   & 59.81($\pm0.12$)               & 59.50($\pm0.05$)               & 59.47($\pm0.21$)               & 59.76($\pm0.12$)          \\ \cline{1-1}
        FitNet            & 58.92($\pm0.20$)               & 58.53($\pm0.49$)               & 58.59($\pm0.07$)               & 58.37($\pm0.11$)          \\ \cline{1-1}
        AT                & 58.52($\pm0.17$)               & 58.74($\pm0.09$)               & 58.60($\pm0.07$)               & 57.87($\pm0.24$)          \\ \cline{1-1}
        PKT               & 58.57($\pm0.17$)               & 58.74($\pm0.05$)               & 58.96($\pm0.27$)               & 58.81($\pm0.06$)          \\ \cline{1-1}
        TA(20)            & -                              & 59.60($\pm0.25$)               & 59.45($\pm0.09$)               & 59.14($\pm0.18$)          \\ \cline{1-1}
        TA(32)            & -                              & -                              & 59.68($\pm0.12$)               & 59.65($\pm0.09$)          \\ \hline
        PrUE($s=20\%$)    & 59.54($\pm0.06$)               & 59.66($\pm0.12$)               & 59.95($\pm0.05$)               & 59.56($\pm0.14$)          \\ \cline{1-1}
        PrUE($s=50\%$)    & \textbf{59.9}($\pm0.30$)       & \textbf{59.71}($\pm0.25$)      & \textbf{60.03}($\pm0.37$)      & 59.85($\pm0.27$)          \\ \cline{1-1}
        PrUE($s=90\%$)    & 58.89($\pm0.32$)               & 59.44($\pm0.19$)               & 59.85($\pm0.20$)               & \textbf{60.17}($\pm0.08$) \\ \hline
    \end{tabular}
\end{table}

\begin{figure}[htbp]
    \centering
    \begin{subfigure}[b]{0.48\linewidth}
        \centering
        \includegraphics[width=0.9\textwidth]{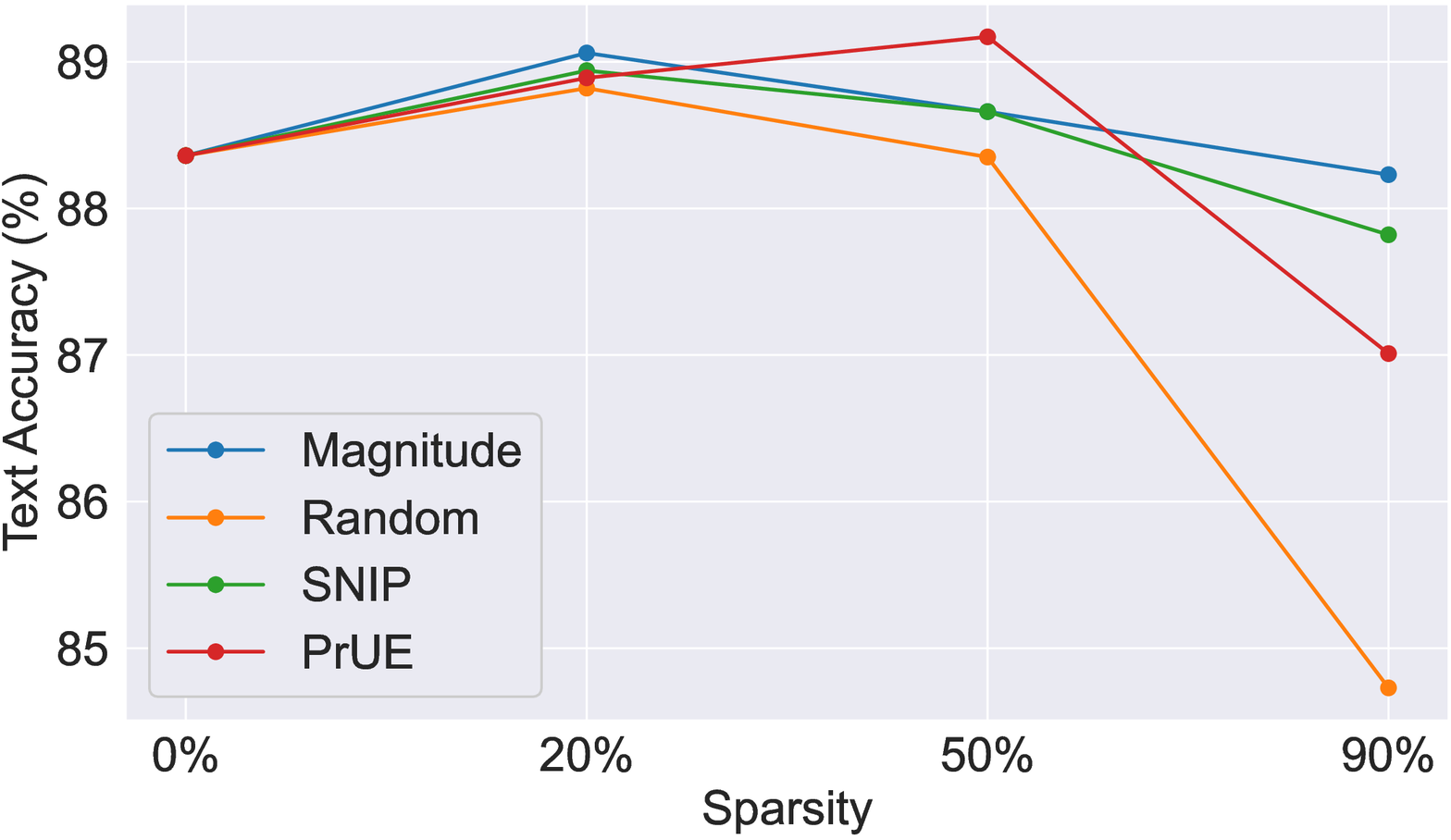}
        \caption{ResNet-20}
    \end{subfigure}
    % \quad
    \begin{subfigure}[b]{0.48\linewidth}
        \centering
        \includegraphics[width=0.9\textwidth]{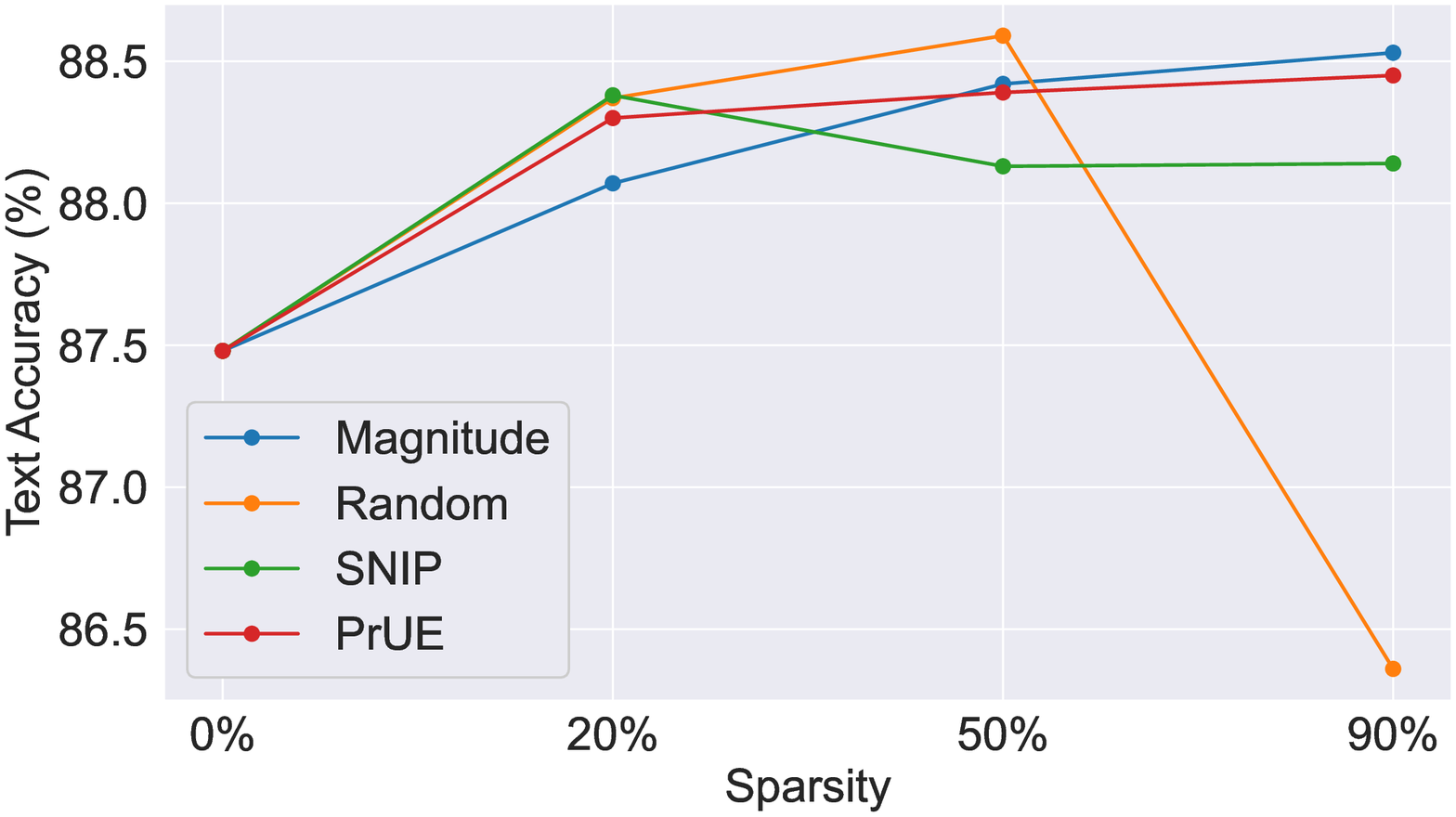}
        \caption{ResNet-32}
    \end{subfigure}
    % \quad
    \begin{subfigure}[b]{0.48\linewidth}
        \centering
        \includegraphics[width=0.9\textwidth]{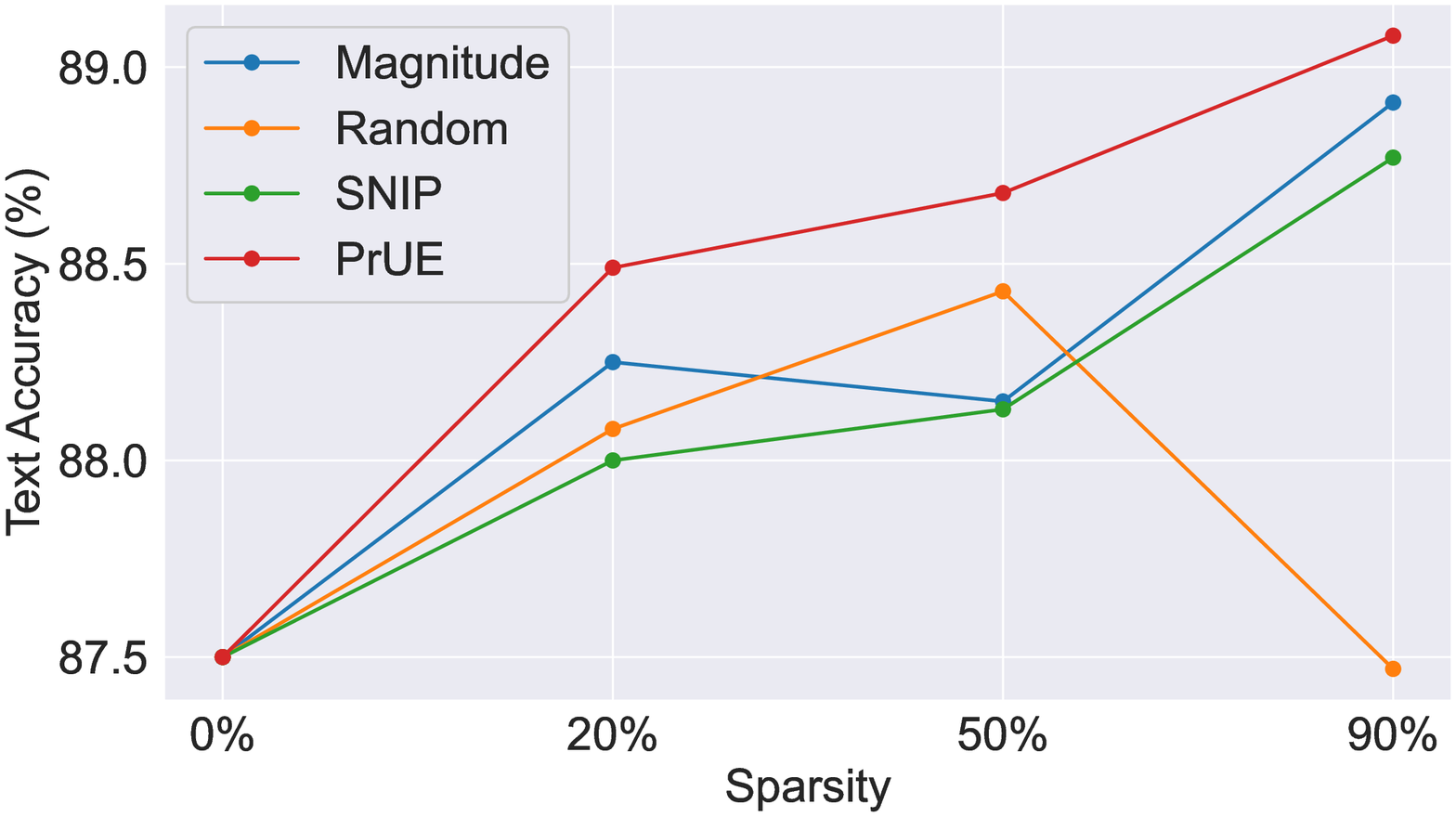}
        \caption{ResNet-56}
    \end{subfigure}
    % \quad
    \begin{subfigure}[b]{0.48\linewidth}
        \centering
        \includegraphics[width=0.9\textwidth]{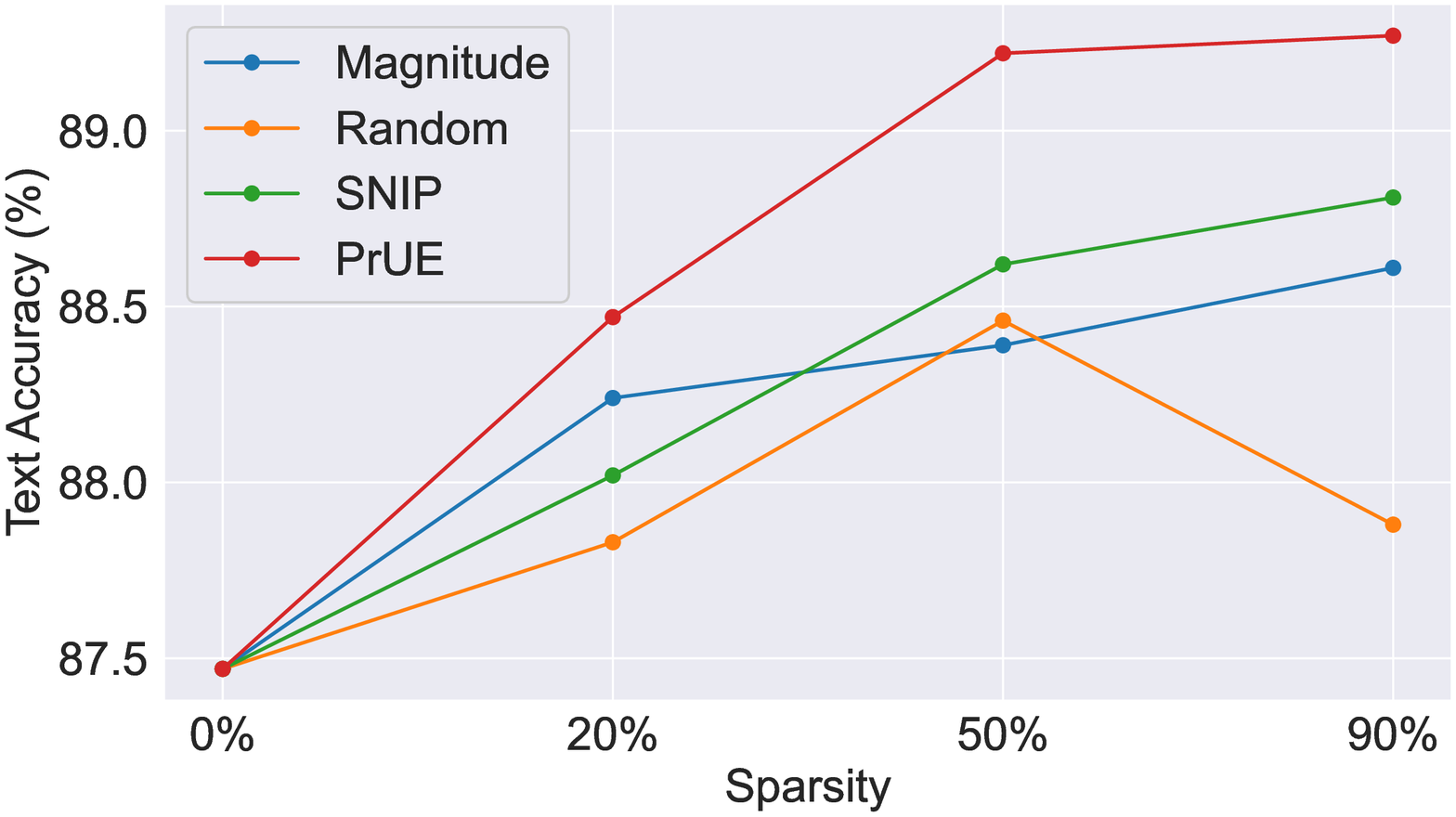}
        \caption{ResNet-110}
    \end{subfigure}
    \caption{Distillation accuracy of sparse teacher networks obtained using different pruning methods.}\label{prune_results}
\end{figure}

As illustrated in Fig.~\ref{prune_results}, our strategy of distilling knowledge from sparse networks can effectively improve the generalization behavior of student networks. Even if the weights in the network are randomly removed, students can still benefit from it. We also notice that PrUE could only exhibit similar performance to other pruning methods on shallower networks. Such as on 90\% sparse ResNet-32, PrUE exhibits lower distillation performance (87.95\%) than Magnitude (88.53\%) and SNIP (88.14\%). But as the network grows, our method achieves better results (up to 89.27\%). This result suggests that while previous work has argued that the large capacity gap between teachers and students results in lower performance gains, our approach allows researchers to break the restriction and use deeper networks to obtain further improve student accuracy.

\subsubsection{The Impact of Sparisty.} We also find that inappropriate sparsity affects the distillation results of all pruning algorithms. For instance, ResNet-20 with 90\% sparsity could face a 1-2\% drop in distillation accuracy, although this result still outperforms traditional distillation methods in Table~\ref{cifar10_results}. While networks with more parameters like ResNet-110 can endure a higher sparsity ratio. Overall, if the size of teachers is much larger than that of students, we suggest a higher sparsity to bridge the capacity gap.

\subsection{Distillation on Large-Scale Datasets}
In this section, we consider practical applications on more challenging datasets. In practice, some large convolutional networks have been proposed to achieve better results on ImageNet tasks. On the other hand, researchers designed some lightweight networks to reduce overhead and accelerate inference. We aim to answer whether PrUE still works between these two different network structures.

\begin{table}[htbp]
    \centering
    \caption{The test accuracy (\%) and uncertainty (1e-2) of sparse teacher networks on Tiny-ImageNet and ImageNet.}
    \label{imagenet_teacher}
    \begin{tabular}{|c|cc|cc|cc|cc|}
        \hline
        \multirow{2}{*}{} & \multicolumn{2}{c|}{$s$=0} & \multicolumn{2}{c|}{$s$=20\%} & \multicolumn{2}{c|}{$s$=50\%} & \multicolumn{2}{c|}{$s$=90\%}                                                                           \\ \cline{2-9}
                          & \multicolumn{1}{c|}{Acc.}  & Uncer.                        & \multicolumn{1}{c|}{Acc.}     & Uncer.                        & \multicolumn{1}{c|}{Acc.} & Uncer. & \multicolumn{1}{c|}{Acc.} & Uncer. \\ \hline
        ResNext-50        & 65.32                      & 1.35                          & 66.24                         & 1.34                          & 68.07                     & 1.29   & 64.09                     & 3.24   \\
        EfficientNet-B2   & 72.32                      & 34.66                         & 72.62                         & 34.70                         & 72.45                     & 34.85  & 71.03                     & 35.14  \\ \hline
    \end{tabular}
\end{table}

\begin{table}[htbp]
    \centering
    \caption{The test accuracy of student network distilled by sparse teachers.}
    \label{imagenet_student}
    \begin{tabular}{|c|c|c|c|c|c|c|}
        \hline
        Teacher         & Student            & Vanilla & $s$=0                      & $s$=20\% & $s$=50\% & $s$=90\% \\ \hline
        ResNext-50      & ShuffleNet V2      & 62.09   & \multicolumn{1}{l|}{63.28} & 63.45    & 64.09    & 64.65    \\
        EfficientNet-B2 & MobileNet V3 Small & 60.85   & 60.88                      & 61.18    & 61.22    & 62.12    \\ \hline
    \end{tabular}
\end{table}

We train ResNext-50 on Tiny-ImageNet as teacher network, while ShuffleNetV2 serves as the student. As for ImageNet, we distill knowledge from EfficientNet-B2 into MobileNetV3. Table~\ref{imagenet_teacher} and Table~\ref{imagenet_student} reports their own accuracy and distillation performance, respectively. Our method manages to improve student generalization on real-world datasets. More interestingly, we observed on Tiny-ImageNet that the accuracy of the student network can sometimes exceed that of the teacher network. We believe this suggests that PrUE can be extended to a wider range of settings. Furthermore, we still lack understanding of knowledge distillation, and our proposed method could be a potential tool to shed light on it.

\section{Conclusion}
In this paper, we provided a data-dependent pruning method called PrUE to soften the network predictions, thereby improving its distillation performance. In particular, we proposed a computationally efficient criterion to estimate the effect of weights on uncertainty, and removed those less-contribution weights. We first showed a positive relationship between the uncertainty of the teacher network and its distillation effect through a visualization scheme. The following empirical experiments suggested that PrUE managed to increase the teacher uncertainty, thereby improving the distillation performance. Extensive experiments showed that our method notably outperformed traditional distillation methods. We also found that our strategy of distilling knowledge from sparse teacher networks could improve the generalization behavior of the student network, but the teacher pruned by PrUE tended to exhibit better performance.

\subsubsection{Acknowledgements} This work is supported by The National Key Research and Development Program of China No. 2020YFE0200500 and National Natural Science Funds of China No. 61902394.

% ---- Bibliography ----
%
% BibTeX users should specify bibliography style 'splncs04'.
% References will then be sorted and formatted in the correct style.
%
\bibliographystyle{splncs04}
% \bibliography{mybibliography}

\end{document}